\documentclass[letterpaper]{article} 
\usepackage{aaai2026}  
\usepackage{times}  
\usepackage{helvet}  
\usepackage{courier}  
\usepackage[hyphens]{url}  
\usepackage{graphicx} 
\usepackage{natbib}  
\usepackage{caption} 
\usepackage{amssymb}
\usepackage{amsmath}
\usepackage{pifont}
\newcommand{\cmark}{\ding{51}}%
\newcommand{\xmark}{\ding{55}}%
\usepackage{algorithm}
\usepackage{algorithmic}

\usepackage{xcolor}
\usepackage{soul}
\sethlcolor{yellow!60}

\newcommand{\answerTODO}[1]{\textcolor{red}{#1}} 

\newcommand{\acro}{\textbf{\texttt{TANDEM}}\xspace}
\newcommand{\std}[1]{\scriptsize{$\pm$#1}}

\usepackage{xspace}
\usepackage{booktabs}
\usepackage{multirow}
\usepackage{makecell}
\usepackage{ragged2e}
\usepackage{afterpage}
\usepackage{newfloat}
\usepackage{listings}
\DeclareCaptionStyle{ruled}{labelfont=normalfont,labelsep=colon,strut=off} 
\floatstyle{ruled}
\newfloat{listing}{tb}{lst}{}
\floatname{listing}{Listing}
\title{\acro: Temporal-Aware Neural Detection for Multimodal Hate Speech}
\author {
    Girish A. Koushik\textsuperscript{\rm 1}\thanks{Corresponding author},
    Helen Treharne\textsuperscript{\rm 2},
    Diptesh Kanojia\textsuperscript{\rm 1}
}
\affiliations {
    \textsuperscript{\rm 1}Nature-Inspired Computing \& Engineering, University of Surrey, Guildford, United Kingdom\\
    \textsuperscript{\rm 2}Surrey Centre for Cyber Security, University of Surrey, Guildford, United Kingdom\\
    g.koushik@surrey.ac.uk, h.treharne@surrey.ac.uk, d.kanojia@surrey.ac.uk
}

\begin{document}

\maketitle

\begin{abstract}
Social media platforms are increasingly dominated by long-form multimodal content, where harmful narratives are constructed through a complex interplay of audio, visual, and textual cues. While automated systems can flag hate speech with high accuracy, they often function as ``black boxes'' that fail to provide the granular, interpretable evidence, such as precise timestamps and target identities, required for effective human-in-the-loop moderation. In this work, we introduce \acro, a unified framework that transforms audio-visual hate detection from a binary classification task into a structured reasoning problem. Our approach employs a novel tandem reinforcement learning strategy where vision-language and audio-language models optimize each other through self-constrained cross-modal context, stabilizing reasoning over extended temporal sequences without requiring dense frame-level supervision. Experiments across three benchmark datasets demonstrate that \acro significantly outperforms zero-shot and context-augmented baselines, achieving $0.73$ F1 in target identification on HateMM (a $\approx30\%$ improvement over state-of-the-art) while maintaining precise temporal grounding. We further observe that while binary detection is robust, differentiating between offensive and hateful content remains challenging in multi-class settings due to inherent label ambiguity and dataset imbalance. More broadly, our findings suggest that structured, interpretable alignment is achievable even in complex multimodal settings, offering a blueprint for the next generation of transparent and actionable online safety moderation tools.


\textbf{Disclaimer}: This paper contains references to material that may be disturbing, hateful, or offensive, reflecting the nature of the task.
\end{abstract}


\section{Introduction}  \label{sec:intro}

Hate speech has become increasingly pervasive on social media platforms, where rich combinations of text, images, audio, and video are routinely used to convey and amplify harmful narratives at scale. While textual hate speech has been extensively studied, the proliferation of rich media on social platforms has dramatically expanded the avenues through which harmful content is propagated. Social media platforms such as YouTube, Facebook, X, and Instagram enable users to combine text, images, audio, and video into a single post, creating multimodal content that eludes traditional unimodal detection systems. The rapid spread of hate content is further exacerbated by algorithmic amplification and echo chamber effects, wherein users are increasingly exposed to content that reinforces existing biases and grievances, potentially normalizing discriminatory rhetoric and contributing to social polarization~\cite{cinelli2021echo,zannettou2018origins}. These dynamics make multimodal hate speech a salient problem for both society and automated content analysis systems.

\begin{figure}[!t]
    \centering
    \includegraphics[width=0.99\linewidth]{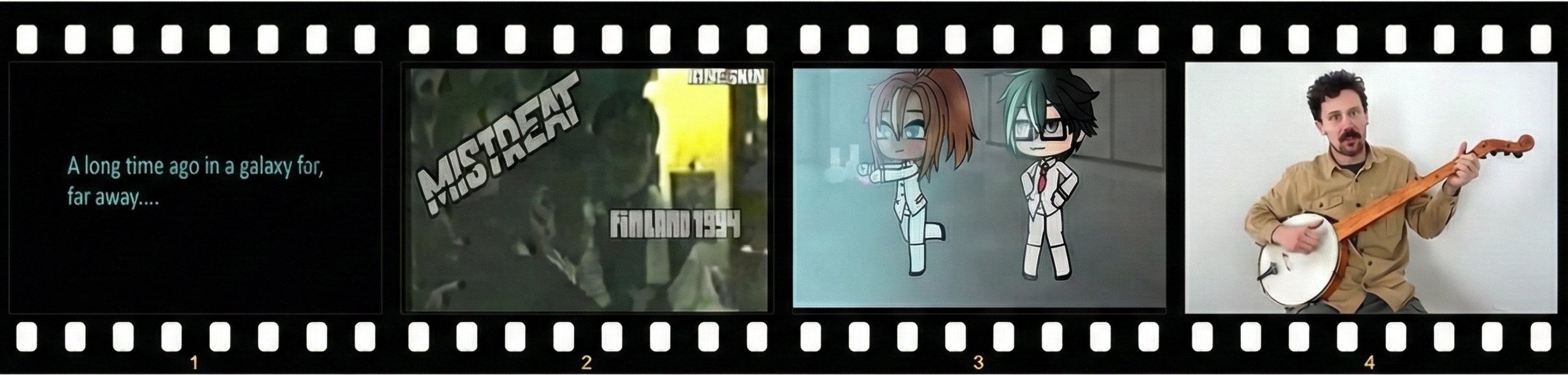}
    \caption{Sample Video Frames from HateMM Dataset~\cite{das2023hatemm}}
    \label{fig:hatemm-sample}
\end{figure}


Detecting and understanding hate speech in audio-visual content is inherently more complex than in text alone. First, the \textit{temporal nature} of videos and speech necessitates effective long sequence processing, requiring models to reason over extended time horizons rather than isolated frames or clips. Second, hate content may only appear briefly within a long video, making \textbf{temporal segmentation} and \textbf{timestamp localization} essential for pinpointing the precise span of hateful behavior. Third, identifying the \textit{target of hate} -- whether a person, community, or a group requires joint reasoning across modalities and careful extraction of semantic cues from visual and auditory streams. Finally, real-world content is rife with noise: background sounds, off-topic visuals, overlapping speech, and low-quality recordings can obscure salient signals and degrade model performance. Existing multimodal detection frameworks often focus on static image-text setups or short clips, leaving long-form audio-visual hate speech largely unaddressed~\cite{hee2024recent}.


Automated hate speech detection systems are critical tools in the toolkit of modern content moderation. Human moderators cannot scale to the volume and velocity of content generated on major platforms, particularly video platforms, where millions of hours of content are uploaded daily. Beyond binary classification (\textit{e.g.}, hate versus non-hate), effective moderation requires \textit{structured outputs} that indicate where within a video hate occurs and who it targets. This enables moderators to rapidly verify and contextualize automated model predictions rather than relying on coarse flags that offer little actionable insight. Policymakers and platform designers increasingly recognize the need for nuanced detection mechanisms that provide interpretable and precise annotations, including temporal boundaries and target labels, to support evidence-based moderation policies and reduce the risk of wrongful takedowns or under-enforcement~\cite{erickson2025content,kentmen2025hate}.

Existing automated moderation approaches largely focus on coarse-grained hate classification, often treating multimodal content as static or short-form inputs. While recent work has explored multimodal fusion and contextual cues, most systems remain limited to video-level labels or short clips, offering little support for temporal localization or explicit target identification~\cite{yue2025multimodal,rehman2025implihatevid,koushik2025towards}. As a result, these approaches provide limited actionable insight for human moderators, who must still manually inspect long videos to verify model predictions. Moreover, current methods rarely address the challenges posed by long-form audio-visual content, cross-modal temporal misalignment, and noisy real-world signals that are prevalent on modern social media platforms.

Motivated by these limitations, we design our approach with online moderation as a first-class objective. Rather than producing a single classification score, our framework explicitly identifies \emph{where} hate occurs in a video and \emph{who} it targets, enabling more efficient verification and intervention. This perspective directly informs the design of \acro, which emphasizes structured outputs, temporal grounding, and cross-modal consistency. We propose \acro as a unified framework for structured multimodal hate understanding in long audio-visual content. Our main contributions are as follows:
\begin{itemize}
    \item We introduce a scalable approach for processing long hateful videos using cross-modal context caching, enabling efficient reasoning over extended audio-visual sequences.
    \item We propose a novel \emph{tandem reinforcement learning} strategy that jointly optimizes vision-language and audio-language models through self-constrained cross-modal context, stabilizing policy optimization and improving cross-modal alignment.
    \item We demonstrate superior structured prediction capabilities, achieving an accuracy of $0.78$ and a target identification F1 score of $0.73$ on HateMM, surpassing strong zero-shot baselines by large margins. We further validate \acro on MultiHateClip and ImpliHateVid, showing robust generalization in multi-class and implicit hate settings.
    \item We provide extensive quantitative and qualitative analyses that highlight both the effectiveness and interpretability of the proposed framework for real-world moderation settings.
\end{itemize}

\section{Related Work}  \label{sec:lit-review}

We review prior work across three pillars central to our approach: existing benchmarks for audio-visual hate speech, methodologies for temporal segmentation in long videos, and recent advances in reinforcement learning for multimodal generation.

\subsection{Audio-Visual Hate Speech}   \label{subsec:av-hs}

The HateMM dataset~\cite{das2023hatemm} represents one of the earliest efforts in audio-visual hate speech detection, comprising over $43$ hours of BitChute videos annotated as either hate or non-hate. Figure~\ref{fig:hatemm-sample} illustrates sample frames from this dataset, showcasing the visual variance involved. For binary classification,~\citet{das2023hatemm} proposed multimodal fusion that fuses video, audio, and speech transcript features, achieving a macro F1 of $0.79$. Subsequent work has largely focused on improving multimodal fusion strategies for the hate versus non-hate task. Beyond the original early-fusion baseline, later studies analyzed modality contributions and fusion designs, showing that simple embedding-level fusion of text, audio, and visual features remains a strong baseline, particularly in video-centric settings~\cite{koushik2025towards}. More recent approaches incorporate explicit temporal modeling, including cross-modal attention and graph-based formulations over video segments, yielding improved performance on HateMM~\cite{yue2025multimodal}. In parallel, evaluations of large pretrained vision-language models (\textit{e.g.}, GPT-4V, LLaVA, Video-LLaMA) in zero-shot or prompting-based settings consistently show inferior performance compared to task-specific multimodal architectures~\cite{wang2024multihateclip,yue2025multimodal}. Finally, contrastive multimodal learning frameworks originally developed for implicit hate detection have demonstrated strong generalization to HateMM, substantially outperforming prior fusion-based methods~\cite{rehman2025implihatevid}.

\begin{table*}[!t]
\centering
\small
\setlength{\tabcolsep}{6pt}
\begin{tabular}{ccccccc}
\toprule
\textbf{Dataset} & \textbf{Labels} & \textbf{Train} & \textbf{Val} & \textbf{Test} & \textbf{Source} & \textbf{Segments / Targets} \\
\midrule
\makecell{HateMM \\ \cite{das2023hatemm}} & Hate / Non-hate & 779 & 87 & 217 & BitChute (EN) & \cmark\ /\ \cmark \\

\makecell{MultiHateClip \\ \cite{wang2024multihateclip}} & Hateful / Offensive / Normal & 1{,}200 & 400 & 400 & YouTube (EN), Bilibili (ZH) & \cmark\ /\ \cmark \\

\makecell{ImpliHateVid \\ \cite{rehman2025implihatevid}} & Explicit / Implicit / Normal & 1{,}009 & 500 & 500 & YouTube (EN) & \xmark\ /\ \xmark \\
\bottomrule
\end{tabular}
\caption{Summary of audio-visual hate speech datasets considered in this work. ``Segments'' denotes time-stamped annotations of hateful content; ``Targets'' denote annotated victim groups. For MHC, we utilize only the English data for experimentation.}
\label{tab:av-hate-datasets}
\end{table*}

Beyond binary labels, HateMM provides fine-grained annotations to support more detailed analysis. Each video includes time-stamped temporal segments identified by annotators as containing hateful content, serving as human rationales for the video-level label. The dataset also specifies the targeted victim group for each hate video. While these annotations enable post hoc analysis of modality contributions, neither the original HateMM study nor subsequent works explicitly train segment-level hate localization or target identification models using this data.

The MultiHateClip (MHC) dataset~\cite{wang2024multihateclip} extends multimodal hate speech detection to a multilingual and fine-grained setting. It contains $2{,}000$ short videos from YouTube (English) and Bilibili (Chinese), annotated into three classes: Hateful, Offensive, and Normal, explicitly distinguishing hate from less severe offensive content. This finer label space substantially increases task difficulty, as models must disambiguate hate from non-hateful but provocative or humorous content. The authors benchmarked a range of state-of-the-art (SoTA) methods for multiclass video hate classification and found that GPT-4V~\cite{gpt4v2023}, when prompted with sampled frames and transcripts, achieved the strongest performance on English videos (macro-F1 $0.63$). However, the study also showed that models frequently confuse hateful content with non-hate offensiveness, a trend we also observe in our experiments. Subsequent work has explored structured multimodal fusion on MHC, including graph-based temporal modeling over video segments with cross-modal interactions, yielding consistent improvements over earlier fusion and attention-based approaches~\cite{yue2025multimodal}. Similar to findings on HateMM, large vision-language models evaluated via prompting exhibit uneven cross-lingual performance and struggle with the rare hateful class, underscoring the need for task-specific multimodal architectures~\cite{wang2024multihateclip,yue2025multimodal}.

Beyond video-level labels, MHC provides detailed annotations for temporal localization and target identification. For each hateful or offensive video, annotators marked the precise temporal segment containing the harmful content and recorded the targeted victim group, with a focus on gender-related categories (\textit{e.g.}, Woman, Man, LGBTQ+, and Other). These annotations enable analysis of when hateful content appears and whom it targets, making MHC a richly annotated hate video dataset. However, the original study did not train dedicated models for segment localization or target prediction, instead using these annotations primarily for dataset analysis and error diagnosis.

The ImpliHateVid (IHV) dataset~\cite{rehman2025implihatevid} targets implicit hate speech in videos, a particularly challenging form of harmful content characterized by veiled, context-dependent expressions that lack explicit slurs or insults. It consists of $2{,}009$ videos annotated into three classes: Non-hate, Implicit Hate, and Explicit Hate, making it the first video dataset to distinguish implicit from explicit hate. Prior datasets such as HateMM and MHC primarily focus on binary or explicit hate labels, limiting their applicability to implicit hate detection. To address this challenge,~\citet{rehman2025implihatevid} proposed a two-stage multimodal contrastive learning framework. In the first stage, modality-specific encoders for text, audio, and visual frames are trained using supervised contrastive loss to learn robust unimodal representations. In the second stage, a cross-modal fusion network further aligns these features into a unified multimodal embedding via contrastive learning. The proposed approach achieves an F1-score of $0.87$ for binary hate detection and a macro-F1 of $0.69$ for three-way classification on ImpliHateVid. Unlike HateMM and MultiHateClip, ImpliHateVid does not include annotations for temporal localization or target identification; we therefore use this dataset solely to evaluate the generalization of our model to explicit and implicit hate classification.

Recent efforts have further advanced fine-grained annotation with datasets like HateClipSeg~\cite{wang2025hateclipseg}, which provides segment-level labels for specific categories (\textit{e.g.}, insult, violence) and explicit target attributes. While valuable for supervised localization, our work focuses on leveraging reinforcement learning to ground these granular signals in long-form video without relying solely on dense frame-level supervision.


\subsection{Long Video Understanding and Temporal Segmentation} \label{subsec:segment-id}

A core capability for long video understanding is temporal localization, identifying when specific events or actions occur. In classic computer vision tasks such as temporal action localization, the goal is to detect and mark the start/end of actions in untrimmed videos. Early approaches often cast this as a sliding-window classification problem: the video is divided into many overlapping segments (windows), each treated as a candidate classifying a certain action. This brute-force strategy was improved by proposal-based methods that generate fewer candidate segments likely to contain actions. For example,~\citet{escorcia2016daps} introduced a deep proposal network to suggest relevant temporal snippets, and~\citet{shou2016temporal} built an end-to-end 3D CNN that directly learns to predict segments. \citet{zhao2017temporal} went further with structured segment networks (SSN) to model the temporal structure within each candidate segment. These fully supervised methods assume precise annotations of action boundaries for training. They work well on shorter videos or curated clips, but scaling to truly long videos (\textit{e.g.}, hours of footage like in the case of HateMM) remains challenging due to the sheer number of windows or proposals needed. 

State-of-the-art multimodal Large Language Models (mLLMs) have recently moved beyond simple proposal networks. New architectures such as TriSense~\cite{li2025watch} explicitly integrate audio, visual, and speech streams for holistic temporal reasoning. Furthermore, modern grounding approaches have adopted specialized mechanisms such as adaptive head-switching and dedicated time-tokens to interleave precise timestamp regression with textual generation~\cite{guo2025vtg,guo2024trace}. \mbox{\citet{sun2025multihateloc}} combined modality-specific temporal encoders with top-$K$ multiple instance learning (MIL) objective to infer hateful segments given video-level labels. They achieve a mean average precision (mAP) of $0.645$ on HateMM and $0.445$ on MHC datasets, respectively. Our framework complements these architectural advances by introducing a training-side innovation -- tandem reinforcement learning, that stabilizes these long-horizon predictions.


\begin{figure*}
    \centering
    \includegraphics[width=0.9\linewidth]{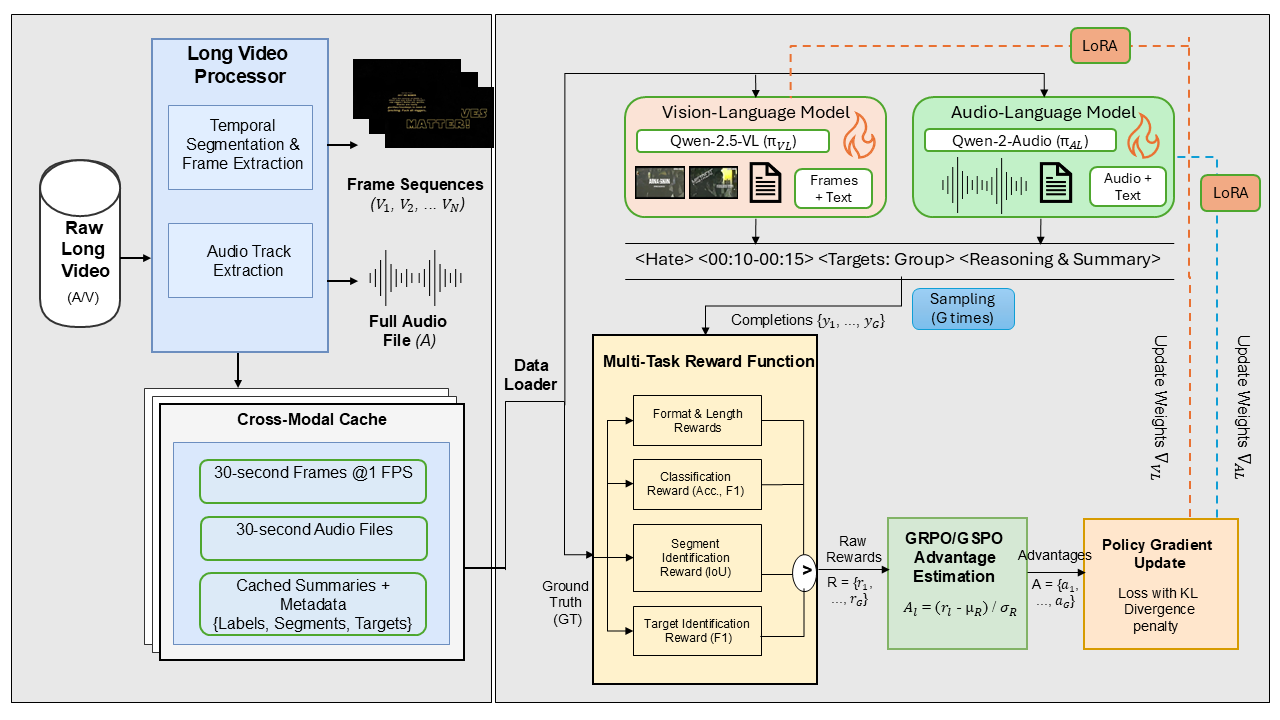}
    \caption{The \acro Architecture. The system employs a tandem reinforcement learning strategy where vision and audio models are iteratively updated while conditioning on each other’s context. The \textit{Long Video Processor} handles segmentation, while the \textit{Multi-Task Reward Function} computes a composite score based on summary length, output formatting, classification accuracy, segment localization, and target identification to guide the policy gradient update.}
    \label{fig:tandem}
\end{figure*}

\subsection{RL for Multimodal Models}   \label{subsec:rl-mm}

More recent multimodal models have begun to explore reinforcement learning (RL) as a mechanism for handling long-horizon reasoning and decision-making in vision-language (VL) and audio-language (AL) tasks, particularly where dense supervision is unavailable or poorly aligned with end objectives. RL has been applied to tasks such as visual question answering, image and video captioning, and multimodal summarization, where models are optimized using sequence-level or outcome-based rewards rather than token-level likelihoods, leading to improved grounding, reduced hallucinations, and better alignment with human judgments~\cite{takada2024direct,li2025self}. A closely related direction is reinforcement learning with verifiable rewards (RLVR), where rewards are derived from deterministic checks rather than subjective preference models~\cite{lambert2024tulu}. While originally studied in text-only reasoning, RLVR has recently been extended to VL tasks through proxy verification signals such as downstream question-answering accuracy conditioned on generated captions, enabling reproducible and fully automatic reward signals for otherwise open-ended multimodal generation~\cite{xing2025caprl}. These developments suggest that RL-based formulations are particularly well suited to temporally extended multimodal reasoning, where models must iteratively select, refine, or validate segments over long contexts.

This ability to precisely localize and reason over salient temporal segments is especially critical in audio-visual hate speech moderation, where harmful content may be sparse, transient, or context-dependent. Accurate identification of hateful segments directly supports efficient moderation by reducing the need for exhaustive review of entire videos. Moreover, target identification is a crucial complementary capability in audio-visual hate speech understanding, as recognizing the target of abuse often requires reasoning over speech, visuals, and their temporal alignment. Despite its importance, to the best of our knowledge, existing audio-visual hate speech datasets and methods do not explicitly address the combined tasks of hate segmentation and target identification in long-form content. Our work directly addresses this gap by framing audio-visual hate understanding as a temporally grounded multimodal reasoning problem.


\section{Methodology}   \label{sec:methodology}

We introduce \acro, a unified framework for structured multimodal hate understanding in long audio-visual content. The following sections outline our pipeline, which encompasses scalable audio-visual processing, structured segment identification, and a novel tandem reinforcement learning strategy for cross-modal optimization.

\subsection{Long Video Processing}  \label{subsec:long-video}

To handle arbitrarily long videos, we partition each input video into manageable 30-second chunks. For each chunk, we extract the audio track and save it as a mono $16 kHz$ WAV file (if no audio is present, a silent placeholder .wav file is used). Simultaneously, we sample key video frames from the 30-second segment to capture its visual content. We employ scene-change detection to split the chunk into shots and select representative keyframes (typically one frame per scene). This approach ensures that dynamic content is well-represented without requiring every frame. If a video segment is too long or scene detection fails, the system falls back to uniform frame sampling (approximately $1$ frame per second, up to a maximum of $24$ frames per chunk). Each chunk is thus characterized by a set of key video frames and a synchronized audio clip. This preprocessing strategy enables our pipeline to scale to long videos by processing one chunk at a time, while preserving both the visual and auditory context of the original content.

\subsection{Hate Segment and Target Identification} \label{hsubsec:segment-id}

Within each $30$-second chunk, our system identifies hate speech segments and their targets using a dual-modal analysis. The chunk’s video frames are analyzed by a VL model, and the audio track by an AL model. Each model produces an XML-formatted output with the following fields: a step-by-step $<reasoning>$ explaining the content, a $<classification>$ of the chunk’s content (\textit{e.g.} Hate vs Non Hate for binary cases, or Normal/Offensive/Hateful for multi-class cases), a $<timestamps>$ field indicating the time span(s) of any detected hateful content, a $<targets>$ field listing the target of hate speech, and a concise $<summary>$ of the chunk. If no hate speech is present, the model outputs a non-hateful classification (\textit{e.g.}, Non Hate or Normal) and uses a placeholder such as ``No hate timestamps'' in the $<timestamps>$ field and ``None'' in $<targets>$. The target identification is constrained to a fixed taxonomy of groups provided in the prompt (for instance, Blacks, Jews, Muslims, LGBTQ, etc., or Christian, Woman, Man, etc., depending on the dataset). This structured prediction allows us to pinpoint where hate occurs in the video and whom it is directed against, with the reasoning and summary providing interpretability. Notably, the VL model’s summary focuses on visual cues (including any on-screen text or symbols), while the AL model’s summary highlights the audible content; together, these help cross-verify the context of any detected hate speech.


\subsection{RL for VL and AL Models}    \label{subsec:RL-vlal}

We fine-tune both the VL and AL models using RL to enable accurate, temporally grounded, and context-aware hate understanding. All RL runs are initialized from models cold-started via supervised fine-tuning (SFT) on a small curated set of structured outputs. We adopt Group Relative Policy Optimization (GRPO)~\cite{shao2024deepseekmath} and its sequence-level variant Group Sequence Policy Optimization (GSPO)~\cite{zheng2025group}, which are well-suited for structured prediction with non-differentiable rewards (as shown in Figure~\ref{fig:tandem}).

A central challenge in multimodal RL is stabilizing learning when each modality depends on the other’s predictions. To address this, we optimize the VL and AL models in \acro through alternating updates. At each phase ($10$ steps), one modality is updated while the other is held fixed, allowing the trainable model to condition on a stable cross-modal signal. Concretely, during a VL update phase, the AL model remains frozen and provides structured audio-derived context; in the subsequent phase, the roles are reversed. This tandem optimization prevents collapse into unimodal shortcuts and promotes consistent cross-modal grounding.

To further stabilize RL, we introduce a \emph{self-constrained context round (SCCR)} prior to policy updates. In SCCR, a frozen modality performs a zero-shot inference pass to generate a structured context that conditions the other modality during RL. Formally, given audio input $x^{a}$, the AL policy produces
\begin{equation}
\tilde{y}^{a} = \pi_{\theta_a}(x^{a}),
\qquad
\pi_{\theta_v}(y \mid x^{v}, \tilde{y}^{a})
\label{eq:sccr}
\end{equation}
where $\tilde{y}^{a}$ is the self-generated audio context injected into the VL policy. The same procedure is applied symmetrically when optimizing the AL model using visual context. SCCR constrains policy updates using model-generated but structured signals, reducing reward sparsity and preventing early-stage drift.

Each model output is a structured prediction
$y = \{\hat{c}, \hat{\tau}, \hat{z}, \hat{s}\}$,
where $\hat{c}$ denotes the hate classification, $\hat{\tau}$ the predicted hate timestamps, $\hat{z}$ the identified targets, and $\hat{s}$ a textual summary. The RL objective is driven by a composite reward that captures semantic correctness, temporal grounding, and structural validity:
\begin{equation}
\begin{split}
R(y, y^\ast) &= \lambda_c \mathbb{CE}[\hat{c} = c^\ast]
+ \lambda_\tau \mathrm{IoU}(\hat{\tau}, \tau^\ast) \\
&\quad + \lambda_z \mathrm{F1}(\hat{z}, z^\ast) 
+ \lambda_f R_{\text{len}}
+ \lambda_f R_{\text{fmt}}
\end{split}
\label{eq:reward}
\end{equation}
where $y^\ast$ denotes ground-truth annotations, $\mathbb{CE}$ denotes cross-entropy loss between predicted and dataset labels, $R_{\text{len}}$ restricts the model's summary length and $R_{\text{fmt}}$ enforces adherence to the required output schema. This formulation subsumes the individual rewards described above for classification, timestamp localization, target identification, and formatting, with weights $\lambda$ controlling their relative influence, treating the task as RLVR~\cite{lambert2024tulu}.

With SCCR conditioning, both modalities are optimized using GRPO or GSPO. For a policy $\pi_\theta$, the optimization objective is
\begin{equation}
\mathcal{L}_{\text{RL}}(\theta) =
- \mathbb{E}_{y \sim \pi_\theta(\cdot \mid x, \tilde{y})}
\left[
\big(R(y) - \bar{R}\big)
\log \pi_\theta(y \mid x, \tilde{y})
\right]
\label{eq:grpo}
\end{equation}
where $\bar{R}$ is a group-wise baseline computed over sampled trajectories, $x$ represents the input audio or video data (\textit{e.g.}, $x^a$ for audio, $x^v$ for video frames), while $y$ represents the structured prediction containing the hate classification, timestamps, target identities, and summary. GRPO computes this baseline over token-level groups, while GSPO operates on full sequence-level samples, improving stability for long structured outputs. During each update, the counterpart modality remains frozen, ensuring that improvements arise from better alignment with cross-modal context rather than co-adaptation.

Finally, when SFT is used, the overall training objective combines supervised and reinforcement signals, with SFT providing structural priors and SCCR-conditioned RL refining temporal localization and target identification. This staged optimization yields stable training dynamics and consistently improves structured multimodal predictions, as demonstrated in Section~\ref{sec:exp-results}.

\begin{table*}[!t]
\centering
\small
\setlength{\tabcolsep}{5.2pt}
\renewcommand{\arraystretch}{1.15}
\resizebox{\linewidth}{!}{
\begin{tabular}{llccccccc}
\toprule
\multicolumn{2}{c}{\textbf{Setting}} &
\multicolumn{3}{c}{\textbf{Classification}} &
\multicolumn{2}{c}{\textbf{Timestamping}} &
\multicolumn{2}{c}{\textbf{Target ID}} \\
\cmidrule(lr){1-2}\cmidrule(lr){3-5}\cmidrule(lr){6-7}\cmidrule(lr){8-9}
\textbf{Method} & \textbf{Model} &
\textbf{Acc} & \textbf{F1 (M)} & \textbf{F1 (W)} &
\textbf{Avg IoU} & \textbf{Acc@0.5} &
\textbf{Avg F1} & \textbf{Exact Match} \\
\midrule

\multicolumn{9}{l}{\textbf{HateMM (Binary)}} \\
\midrule
ZS & Gemini-2.5-Flash (A+V) & 0.77 & 0.77 & -- & 0.46 & 0.47 & \textbf{0.42} & \textbf{0.33} \\
ZS & Qwen2.5-VL-7B (V) & 0.71 & 0.54 & -- & 0.09 & 0.04 & 0.15 & 0.13 \\
ZS & Qwen2-Audio-7B (A) & 0.56 & 0.57 & -- & 0.00 & 0.00 & 0.03 & 0.00 \\
ZS & Qwen3-Omni-30B-A3B-Thinking (A+V) & 0.40 & 0.57 & -- & \textbf{0.53} & \textbf{0.55} & 0.41 & 0.32 \\
\cmidrule(lr){1-9}
Ctx-Aug ZS & Qwen2.5-VL-7B (V + audio transcript) & 0.72 & 0.60 & -- & 0.19 & 0.16 & 0.21 & 0.17 \\
Ctx-Aug ZS & Qwen2.5-VL-7B (V + A context) & 0.67 & 0.56 & -- & 0.10 & 0.05 & 0.13 & 0.08 \\
\midrule
& \citet{yue2025multimodal} & 0.82 & 0.77 & -- & -- & -- & -- & -- \\
& \citet{koushik2025towards} & 0.85 & 0.85 & -- & -- & -- & -- & -- \\
& \citet{rehman2025implihatevid} & \textbf{0.97} & \textbf{0.97} & -- & -- & -- & -- & -- \\
\midrule
\multicolumn{9}{l}{\textbf{MultiHateClip (MHC, Multiclass)}} \\
\midrule
ZS & Gemini-2.5-Flash (A+V) & 0.61 & \textbf{0.36} & \textbf{0.56} & 0.07 & 0.06 & \textbf{0.29} & \textbf{0.27} \\
ZS & Qwen2.5-VL-7B (V) & 0.58 & 0.33 & 0.55 & 0.11 & 0.09 & 0.25 & 0.21 \\
ZS & Qwen2-Audio-7B (A) & \textbf{0.67} & 0.27 & 0.53 & 0.07 & 0.07 & 0.18 & 0.18 \\
ZS & Qwen3-Omni-30B-A3B-Thinking (A+V) & 0.64 & 0.28 & 0.54 & \textbf{0.13} & \textbf{0.15} & 0.25 & 0.24 \\
\midrule
& \citet{wang2024multihateclip} & -- & 0.54 & -- & -- & -- & -- & -- \\
\midrule

\multicolumn{9}{l}{\textbf{ImpliHateVid (IHV, Multiclass)}} \\
\midrule
ZS & Qwen3-Omni-30B-A3B-Thinking (A+V) & 0.62 & 0.46 & 0.52 & -- & -- & -- & -- \\
\midrule
& \citet{rehman2025implihatevid} & -- & \textbf{0.69} & -- & -- & -- & -- & -- \\
\bottomrule
\end{tabular}
}
\caption{\textbf{Zero-shot (ZS), and Baseline results.} We report classification (Acc, Macro F1, Weighted F1), timestamping (Avg IoU, Acc@0.5), and target identification (Avg F1, Exact Match). All values are higher-better. Best per dataset/column is in bold. V $\rightarrow$ Video, A $\rightarrow$ Audio.}
\label{tab:zs_context_results}
\end{table*}

\subsection{Experimental Setup} \label{subsec:exp-setup}

We evaluate our approach on two multimodal hate speech datasets: HateMM and MHC datasets for the three tasks outlined in Section~\ref{sec:intro}. Further, we evaluate our best performing approach on the IHV dataset for the classification task only (explicit/implicit/no hate) for generalizability (as shown in Table~\ref{tab:av-hate-datasets}). We utilize the pre-trained Qwen2.5-VL-7B-Instruct~\cite{bai2025qwen2} and Qwen2-Audio-7B-Instruct~\cite{chu2024qwen2} as our base models. To ensure efficiency, we employ Low-Rank Adaptation (LoRA)~\cite{hu2022lora} on transformer layers rather than full fine-tuning. Training occurs in two stages: first, Supervised Fine-Tuning (SFT) on a curated subset of $100$ high-quality videos filtered via Qwen3-Omni-30B-A3B-Thinking~\cite{xu2025qwen3omnitechnicalreport} predictions; second, the proposed tandem RL procedure (GRPO/GSPO) applied to the full training sets.

\paragraph{Evaluation Metrics}
For evaluation, models process test videos in $30$-second chunks. We aggregate chunk-level predictions to video-level labels for final scoring; metrics for timestamps and targets are computed exclusively on positive (hateful) instances to prevent unusually high scores. We experimented with this and found that when we use negative (non-hateful) instances, the evaluation treats ``no timestamps'' and ``no targets'' as correct and increases the scores. The full set (positive \& negative instances) numbers will be inflated as normal samples, with empty timestamps and empty target labels counted as perfect matches (shown in Table~\mbox{\ref{tab:pos_neg}}). 

We benchmark against strong zero-shot baselines, including Gemini-2.5-Flash~\cite{comanici2025gemini} and Qwen3-Omni. We assess performance across three dimensions:
\begin{itemize}
    \item \textbf{Classification}: We report Accuracy Macro-F1. For MHC and IHV, we also report Weighted-F1.
    \item \textbf{Temporal Localization}: We measure average Intersection-over-Union (Avg IoU) between predicted and ground-truth intervals, alongside Accuracy@0.5 (proportion of predictions with $IoU > 0.5$).
    \item \textbf{Target Identification}: We use the average F1 score across target labels and Exact Match Accuracy, which requires the predicted set of targets to differ in no way from the ground truth.
\end{itemize}

Detailed training hyperparameters, hardware specifications, and SFT filtering protocols are provided in Appendix~\ref{sec:app-imp}.

\begin{table*}[!t]
\centering
\small
\setlength{\tabcolsep}{4pt}
\renewcommand{\arraystretch}{1.12}
\resizebox{\linewidth}{!}{
\begin{tabular}{llccccccccc}
\toprule
\multicolumn{3}{c}{\textbf{Setting}} &
\multicolumn{3}{c}{\textbf{Classification}} &
\multicolumn{2}{c}{\textbf{Timestamping}} &
\multicolumn{2}{c}{\textbf{Target ID}} \\
\cmidrule(lr){1-3}\cmidrule(lr){4-6}\cmidrule(lr){7-8}\cmidrule(lr){9-10}
\textbf{Method} & \textbf{Model} & \textbf{Cfg} &
\textbf{Acc} & \textbf{F1 (M)} & \textbf{F1 (W)} &
\textbf{Avg IoU} & \textbf{Acc@0.5} &
\textbf{Avg F1} & \textbf{Exact Match} \\
\midrule

\multicolumn{10}{l}{\textbf{HateMM (Binary)}}\\
\midrule
SFT with Qwen-Omni data & Qwen2.5-VL-7B (V+A) & -- & 0.73 \std{0.08} & 0.62 \std{0.05} & -- & -- & -- & 0.23 \std{0.03} & 0.17 \std{0.04} 
\\
(100 videos only) & Qwen2-Audio-7B (A+V) & -- & \textbf{0.78} \std{0.08} & \textbf{0.79} \std{0.06} & -- & 0.18 \std{0.03} & 0.11 \std{0.02} & 0.71 \std{0.11} & 0.58 \std{0.13} 
\\
SCCR + GSPO & Qwen2.5-VL-7B (V+A) & a & 0.73 \std{0.08} & 0.59 \std{0.05} & -- & -- & -- & 0.29 \std{0.04} & 0.26 \std{0.06} 
\\
 & Qwen2-Audio-7B (A+V) & -- & 0.71 \std{0.07} & 0.70 \std{0.06} & -- & 0.32 \std{0.06} & 0.29 \std{0.05} & 0.29 \std{0.04} & 0.16 \std{0.04} 
\\
SFT + SCCR + GSPO & Qwen2.5-VL-7B (V+A) & a & 0.78 \std{0.08} & 0.73 \std{0.06} & -- & -- & -- & 0.55 \std{0.08} & 0.48 \std{0.11} 
\\
 & Qwen2-Audio-7B (A+V) & -- & 0.77 \std{0.08} & 0.78 \std{0.06} & -- & 0.18 \std{0.03} & 0.08 \std{0.01} & 0.73 \std{0.11} & 0.57 \std{0.13} 
\\
SFT + SCCR + GRPO & Qwen2.5-VL-7B (V+A) & b & 0.71 \std{0.07} & 0.54 \std{0.04} & -- & -- & -- & 0.28 \std{0.04} & 0.23 \std{0.05} 
\\
 & Qwen2-Audio-7B (A+V) & -- & 0.75 \std{0.08} & 0.76 \std{0.06} & -- & \textbf{0.43} \std{0.08} & \textbf{0.31} \std{0.06} & \textbf{0.73} \std{0.11} & \textbf{0.59} \std{0.13} 
\\
SFT + SCCR + GSPO & Qwen2.5-VL-7B (V+A) & c & 0.72 \std{0.08} & 0.59 \std{0.05} & -- & -- & -- & 0.33 \std{0.05} & 0.28 \std{0.06} 
\\
 & Qwen2-Audio-7B (A+V) & -- & 0.75 \std{0.08} & 0.73 \std{0.06} & -- & 0.19 \std{0.03} & 0.11 \std{0.02} & 0.31 \std{0.05} & 0.15 \std{0.03} 
\\
\midrule

\multicolumn{10}{l}{\textbf{MultiHateClip (MHC-en, Multiclass)}}\\
\midrule
SFT with Qwen-Omni data & Qwen2.5-VL-7B (V+A) & -- & 0.59 \std{0.06} & 0.38 \std{0.03} & 0.57 \std{0.05} & 0.13 \std{0.02} & 0.09 \std{0.02} & 0.29 \std{0.04} & 0.27 \std{0.06} 
\\
(MHC 100 videos) & Qwen2-Audio-7B (A+V) & -- & 0.64 \std{0.07} & 0.33 \std{0.03} & 0.55 \std{0.04} & 0.07 \std{0.01} & 0.07 \std{0.01} & 0.19 \std{0.03} & 0.18 \std{0.04} 
\\
SCCR + GRPO (on HateMM) & Qwen2.5-VL-7B (V+A) & b & 0.58 \std{0.06} & 0.33 \std{0.03} & 0.55 \std{0.04} & 0.11 \std{0.02} & 0.09 \std{0.02} & 0.25 \std{0.04} & 0.21 \std{0.05} 
\\
 & Qwen2-Audio-7B (A+V) & -- & 0.64 \std{0.07} & 0.27 \std{0.02} & 0.53 \std{0.04} & 0.07 \std{0.01} & 0.07 \std{0.01} & 0.18 \std{0.03} & 0.18 \std{0.04} 
\\
SCCR + GSPO (on MHC) & Qwen2.5-VL-7B (V+A) & c & 0.59 \std{0.06} & 0.35 \std{0.03} & 0.57 \std{0.05} & 0.13 \std{0.02} & 0.09 \std{0.02} & 0.28 \std{0.04} & 0.25 \std{0.06} 
\\
 & Qwen2-Audio-7B (A+V) & -- & 0.63 \std{0.07} & 0.28 \std{0.02} & 0.53 \std{0.04} & 0.08 \std{0.01} & 0.07 \std{0.01} & 0.22 \std{0.03} & 0.21 \std{0.05} 
\\
SFT + SCCR + GSPO & Qwen2.5-VL-7B (V+A) & a & 0.59 \std{0.06} & \textbf{0.38} \std{0.03} & \textbf{0.57} \std{0.05} & \textbf{0.13} \std{0.02} & \textbf{0.09} \std{0.02} & \textbf{0.29} \std{0.04} & \textbf{0.27} \std{0.06} 
\\
 & Qwen2-Audio-7B (A+V) & -- & \textbf{0.67} \std{0.07} & 0.32 \std{0.03} & 0.54 \std{0.04} & 0.07 \std{0.01} & 0.07 \std{0.01} & 0.19 \std{0.03} & 0.18 \std{0.04} 
\\
\midrule

\multicolumn{10}{l}{\textbf{ImpliHateVid (IHV, Multiclass)}}\\
\midrule
SFT + SCCR + GRPO & Qwen2.5-VL-7B (V+A) & b & 0.64 \std{0.07} & 0.54 \std{0.04} & 0.59 \std{0.05} & -- & -- & -- & -- 
\\
(on HateMM) & Qwen2-Audio-7B (A+V) & -- & 0.61 \std{0.06} & 0.46 \std{0.04} & 0.53 \std{0.04} & -- & -- & -- & -- 
\\
SFT + SCCR + GSPO & Qwen2.5-VL-7B (V+A) & c & 0.64 \std{0.07} & \textbf{0.54} \std{0.04} & 0.59 \std{0.05} & -- & -- & -- & -- 
\\
(on HateMM) & Qwen2-Audio-7B (A+V) & -- & 0.54 \std{0.06} & 0.47 \std{0.04} & 0.53 \std{0.04} & -- & -- & -- & -- 
\\
SFT + SCCR + GSPO & Qwen2.5-VL-7B (V+A) & a & \textbf{0.64} \std{0.07} & 0.54 \std{0.04} & \textbf{0.59} \std{0.05} & -- & -- & -- & -- 
\\
(on MHC) & Qwen2-Audio-7B (A+V) & -- & 0.60 \std{0.06} & 0.48 \std{0.04} & 0.55 \std{0.04} & -- & -- & -- & -- 
\\
\bottomrule
\end{tabular}}

\caption{\textbf{\acro training results with ablations and error bars estimated across datasets.}
We evaluate classification (Acc, Macro F1, Weighted F1), timestamp localization (Avg IoU, Acc@0.5), and target identification (Avg F1, Exact Match). \textbf{SCCR} (\emph{self-constrained context round}) denotes a zero-shot inference pass that produces structured cross-modal context.
For RL runs, the \textbf{Cfg} column indexes the exact sampling/reward configuration. For the VL model, we omit the temporal segmentation as it degraded the classification and target identification performance. However, for MHC, we enable temporal segmentation for the VL model, as it performed better than the AL model. Higher is better for all metrics. Best per dataset/column is in bold and error bars are estimated based on 3 random seed runs ($s \in \{42, 108, 420\}$).}
\label{tab:tandem_all_datasets}
\vspace{2mm}

{\footnotesize
\textbf{Cfg legend:}
(a) sequence-level, $G{=}4$, $w{=}[1,1,1,1,1]$;\,
(b) token-level, $G{=}4$, $w{=}[1,1,1,1,1]$;\,
(c) sequence-level, $G{=}4$, $w{=}[0.15,0.15,0.3,0.2,0.2]$; where $w = [\lambda_{\text{len}}, \lambda_{\text{fmt}}, \lambda_c, \lambda_\tau, \lambda_z]$.
}
\end{table*}

\section{Experimental Results}  \label{sec:exp-results}

We evaluate \acro against strong zero-shot and context-augmented baselines across three hate speech datasets. Our analysis details quantitative performance on classification, temporal localization, and target identification, followed by a qualitative examination of the model's structured reasoning.

\subsection{Baseline Performance}  \label{subsec:baselines}

On HateMM, Gemini-2.5-Flash is the strongest zero-shot baseline for classification and target ID, while Qwen3-Omni-30B-A3B-Thinking model is notably better at temporal localization (higher Avg IoU) despite weaker classification. The two context-augmented variants improve over plain V-only baselines but do not surpass Gemini on overall detection quality, and their temporal localization remains behind the omni model. Recent baselines on HateMM have performed significantly well (as shown in Table~\ref{tab:zs_context_results}); however, they have focused solely on the classification task. On MHC, Gemini again leads on macro-F1, weighted-F1, and target metrics, whereas Qwen3-Omni provides the best timestamp localization (Avg IoU and Acc@0.5) with the baseline~\citet{wang2024multihateclip} performing quite higher in terms of classification macro-F1. For IHV, Qwen3-Omni performs slightly lower than the~\citet{rehman2025implihatevid} baseline.

Gemini-2.5-Flash is the best all-rounder (especially macro-F1 and targets), while the Qwen3-Omni model shines on localization (Avg IoU and Acc@0.5), suggesting they are better at ``where is the hate'' than ``is it hate'' given once it identifies the content as hateful. The context-augmented VL variants are a meaningful step up over plain V-only baselines on HateMM (especially targets), but still lag behind the strongest A+V model for end-to-end performance.

Table~\mbox{\ref{tab:zs_context_results}} also includes strong dataset-specific supervised baselines from prior literature [including \mbox{\cite{yue2025multimodal}}, \mbox{\cite{koushik2025towards}}, \mbox{\cite{rehman2025implihatevid}}], which are primarily optimized for label prediction. These methods remain competitive for classification-only reporting, but they do not provide the same structured outputs for temporal localization and target identification that are required for actionable moderation.


\subsection{\acro}  \label{subsec:tandem-results}

We next evaluate \acro under three training regimes: SFT-only, RL-only (GRPO or GSPO depending on whether sampling is token-level or sequence-level), and SFT followed by RL. To initialize RL with stable cross-modal signals, we optionally run a self-constrained context round (SCCR), a zero-shot inference pass that produces structured context consumed by the opposite modality during optimization. We report task metrics for hate classification, timestamp localization, and target identification when available.

Table~\ref{tab:tandem_all_datasets} presents the performance of TANDEM with estimated error bars. We observe that performance overlaps between different RL configurations are often within the margin of error, though the best configurations consistently outperform baselines. Across datasets, SFT improves overall classification stability, while SFT + SCCR + RL (GRPO) yields the strongest aggregate behavior, particularly on timestamp localization and target identification in HateMM. We note that while error margins indicate some variance across seeds, the SFT + SCCR + GRPO configuration reliably achieves high target identification ($0.73$ F1). RL-only run without SFT remains sensitive to reward/sampling configuration and often trades off between localization and classification. On MHC-en, the best \acro configuration improves macro-F1 and yields the best localization (Avg IoU), indicating that SFT with SCCR-conditioned RL (GSPO) is most beneficial when the dataset demands both multi-class distinction and temporal grounding. On IHV, SCCR-conditioned \acro transfers well from MHC training (since both datasets are multi-class tasks), improving macro-F1 over the zero-shot baseline without any additional training.

The baselines that outperform \mbox{\acro} on HateMM binary classification \textit{i.e.}, \mbox{\cite{yue2025multimodal}}, \mbox{\cite{koushik2025towards}}, \mbox{\cite{rehman2025implihatevid}} optimize only the video-level hate/non-hate decision. This is a strict coarsening of our objective, which jointly predicts the label, temporal hate span(s), and target group(s). As formalized in Appendix~\mbox{\ref{app:coarsening}}, binary classification-only models operate over a substantially smaller hypothesis space, whereas our joint formulation must reason over a combinatorial space of span--target configurations that scales as \mbox{$\mathcal{O}(M T^2)$} with video length. Consequently, a model can achieve high classification accuracy without preserving the boundary-level or target-specific information required for actionable moderation, while the reverse is not true. Therefore, classification-only baselines are informative but not comparable upper bounds for \mbox{\acro}, and the observed trade-off reflects the increased complexity of producing fine-grained, structured predictions.

\afterpage{\afterpage{\afterpage{\afterpage{
\begin{table}[!t]
\centering
\setlength{\tabcolsep}{5pt}
\renewcommand{\arraystretch}{1.12}
\resizebox{\linewidth}{!}{
\begin{tabular}{llcccc}
\toprule
\multicolumn{3}{c}{\textbf{Setting}} &
\multicolumn{3}{c}{\textbf{Binary Classification}} \\
\cmidrule(lr){1-3}\cmidrule(lr){4-6}
\textbf{Method} & \textbf{Model} & \textbf{Cfg} &
\textbf{Acc} & \textbf{F1 (M)} & \textbf{F1 (W)} \\
\midrule
SFT with Qwen-Omni & Q2-Audio-7B & -- & 0.65 \std{0.07} & 0.46 \std{0.04} & 0.57 \std{0.05} 
\\
data (MHC 100 videos) & Q2.5-VL-7B & -- & 0.63 \std{0.07} & 0.57 \std{0.05} & 0.62 \std{0.05} 
\\
\midrule
SCCR + GRPO & Q2.5-VL-7B & a & 0.64 \std{0.07} & \textbf{0.58} \std{0.05} & \textbf{0.63} \std{0.05} 
\\
(trained on MHC) & Q2-Audio-7B & -- & \textbf{0.65} \std{0.07} & 0.47 \std{0.04} & 0.57 \std{0.05} 
\\
\midrule
SFT + SCCR + GRPO & Q2.5-VL-7B & c & 0.63 \std{0.07} & 0.57 \std{0.05} & 0.62 \std{0.05} 
\\
 & Q2-Audio-7B & -- & 0.65 \std{0.07} & 0.45 \std{0.04} & 0.56 \std{0.04} 
\\
\bottomrule
\end{tabular}}
\captionsetup{font=footnotesize}
\caption{\textbf{Binary (Offensive/Normal) classification results on MultiHateClip.} We report Accuracy, Macro F1, and Weighted F1. The \textbf{Cfg} column indexes the sampling configuration (consistent with Table~\ref{tab:tandem_all_datasets}). Error bars are estimated.}
\label{tab:mhc_binary}
\end{table}
}}}}


\subsection{Discussion} \label{subsec:discussion}

Tables~\ref{tab:zs_context_results} and~\ref{tab:tandem_all_datasets} reveal a consistent pattern across all three datasets. While strong zero-shot multimodal models achieve competitive hate classification accuracy, they remain fundamentally limited in structured prediction tasks such as temporal localization and target identification. \acro addresses these limitations by explicitly coupling modalities and optimizing them with task-aligned rewards.

From Table~\ref{tab:zs_context_results}, Gemini-2.5-Flash performs strongly on hate classification, particularly on HateMM, but these gains do not consistently transfer to timestamping or target identification. In contrast, the Qwen3-Omni model achieves substantially higher Avg IoU on HateMM and MHC, indicating stronger temporal grounding, albeit with weaker or less stable classification and target predictions. Context-augmented zero-shot variants improve over unimodal baselines, especially for target identification, but remain clearly behind fully-trained approaches.

Table~\ref{tab:tandem_all_datasets} shows that \acro substantially narrows this gap. Supervised fine-tuning alone yields clear gains in classification and target identification, confirming that structured supervision stabilizes output semantics. However, SFT-only models show limited improvement in timestamp localization, suggesting that temporal grounding is not easily learned from supervised signals alone. Reinforcement learning without SFT further exposes this limitation: RL-only variants can improve localization under certain configurations but often degrade classification or target accuracy, reflecting sensitivity to reward noise and unstable policy updates. The error bars in Table~\ref{tab:tandem_all_datasets} further highlight the volatility of RL optimization in multimodal settings; while mean performance improves, the variance across seeds indicates that reward sensitivity remains a challenge for future work.

\afterpage{\afterpage{\afterpage{
\begin{table}[!t]

\centering
\renewcommand{\arraystretch}{1.12}
\resizebox{\linewidth}{!}{
\begin{tabular}{llccccc}
\hline
\textbf{Dataset} & \textbf{Eval Method} & \multicolumn{3}{c}{\textbf{Classification}} & \textbf{Timestamping} & \textbf{Target ID} \\
 &  & Acc & F1 (M) & F1 (W) & Avg IoU & Avg F1 \\
\hline
\multirow{2}{*}{HateMM}
& \makecell[l]{Positive\\(hateful)} & 0.78 & 0.79 & -- & 0.43 & 0.73 \\
& \makecell[l]{Positive +\\negative} & 0.78 & 0.79 & -- & \textbf{0.52} & \textbf{0.77} \\
\hline
\multirow{2}{*}{MHC-en}
& \makecell[l]{Positive\\(hateful/offensive)} & 0.67 & 0.38 & 0.57 & 0.13 & 0.29 \\
& \makecell[l]{Positive +\\negative} & 0.67 & 0.38 & 0.57 & \textbf{0.64} & \textbf{0.69} \\
\hline
\end{tabular}}
\captionsetup{font=footnotesize}
\caption{Score comparisons for Timestampping and Target identification when using only positive vs positive \& negative (full set) instances.}
\label{tab:pos_neg}

\end{table}
}}}

The strongest and most consistent results are obtained with \textbf{SFT + SCCR + RL}. Across HateMM and MHC, this configuration outperforms both zero-shot baselines (Table~\ref{tab:zs_context_results}) and single-stage training (Table~\ref{tab:tandem_all_datasets}) on nearly all structured metrics. On HateMM, it achieves the highest Avg F1 and Exact Match accuracy for target identification while also improving timestamp localization relative to SFT-only and zero-shot models. On MHC, SFT provides a strong semantic prior, while SCCR-conditioned RL improves temporal grounding, yielding the best macro/weighted F1 and localization scores with minimal loss in overall accuracy. This indicates that \acro effectively balances competing objectives across tasks.

As shown in Table~\mbox{\ref{tab:pos_neg}}, the full set numbers are inflated by normal (non-hateful) samples, where empty timestamps and empty target labels are counted as perfect matches. If the model points to the wrong segment in a hateful video or hallucinates on a clean video, the evaluation metrics decrease, so there is no scope for ``False Positive Grounding.'' Further, the jump in MHC score is significant because the dataset is highly imbalanced, with more than 60\% videos labeled as ``Normal.''

Ablation results further clarify the role of SCCR. Models trained with SCCR consistently outperform their counterparts trained without it, particularly under RL. This confirms that SCCR provides a stable cross-modal signal that mitigates reward sparsity and prevents modality drift during policy optimization. While sequence-level sampling (GSPO) improves stability for MHC, token-level GRPO performs better on HateMM, suggesting that optimal sampling granularity depends on dataset characteristics. Nevertheless, the dominant factor remains the presence of structured context and prior supervision rather than the specific RL variant.

Additional insights are provided by the binary MHC results in Table~\ref{tab:mhc_binary}. When collapsing labels into a binary setting, SFT + SCCR + RL yields consistent improvements in macro F1 ($0.57\pm0.05$ for VL) over both SFT-only and RL-only baselines, indicating that the benefits of tandem optimization extend beyond fine-grained multi-class distinctions. This also suggests that some classification errors in the multiclass setting stem from label ambiguity rather than a failure to detect harmful content.

Finally, results on ImpliHateVid demonstrate that \acro generalizes beyond the training distributions. Despite being trained on HateMM or MHC, SFT + SCCR + RL improves macro-F1 over zero-shot baselines, while trailing a fully supervised in-domain model~\citep{rehman2025implihatevid}. This highlights both the transferability of learned cross-modal alignment and the remaining gap when domain-specific supervision is unavailable.

Taken together, these results support three conclusions. First, zero-shot or context-augmented inference alone is insufficient for structured multimodal hate understanding. Second, reinforcement learning is effective only when grounded by structured supervision and constrained cross-modal context. Third, \acro’s combination of SFT, SCCR, and RL yields consistent gains across classification, localization, and target identification, validating its design as a unified framework rather than a collection of independent components.

To position our method against a dedicated localization baseline, Table~\mbox{\ref{tab:multihateloc_compare}} compares \mbox{\acro} with MultiHateLoc~\mbox{\cite{sun2025multihateloc}} on HateMM and MHC. \mbox{\acro} improves mAP on both datasets and additionally provides strong classification and target-identification performance, highlighting the benefit of a unified multi-task framework over localization-only baselines.

\begin{table}[!t]
\centering

\setlength{\tabcolsep}{4pt}
\renewcommand{\arraystretch}{1.1}
\resizebox{\linewidth}{!}{
\begin{tabular}{llccccc}
\toprule
\multirow{2}{*}{\textbf{Dataset}} & \multirow{2}{*}{\textbf{Method}} & \multicolumn{3}{c}{\textbf{Classification}} & \textbf{Timestamping} & \textbf{Target ID} \\
\cmidrule(lr){3-5}\cmidrule(lr){6-6}\cmidrule(lr){7-7}
 &  & \textbf{Acc} & \textbf{F1 (M)} & \textbf{F1 (W)} & \textbf{mAP} & \textbf{Avg F1} \\
\midrule
\multirow{2}{*}{HateMM}
& MultiHateLoc & -- & -- & -- & 0.645 & -- \\
& \acro (ours) & 0.78 & 0.79 & -- & \textbf{0.71} & 0.77 \\
\midrule
\multirow{2}{*}{MHC-en}
& MultiHateLoc & -- & -- & -- & 0.445 & -- \\
& \acro (ours) & 0.67 & 0.38 & 0.57 & \textbf{0.62} & 0.69 \\
\bottomrule
\end{tabular}}
\captionsetup{font=footnotesize}
\caption{Comparison with MultiHateLoc on localization-focused evaluation. While MultiHateLoc reports only localization (mAP), \acro jointly supports classification, timestamping, and target identification.}
\label{tab:multihateloc_compare}

\end{table}

\begin{table*}[!t]
\centering
\small
\renewcommand{\arraystretch}{1.3}
\newcolumntype{L}[1]{>{\RaggedRight\arraybackslash}p{#1}}
\resizebox{\linewidth}{!}{
\begin{tabular}{l l L{5cm} c c c l}
\toprule
\textbf{Dataset} & \textbf{Video ID} & \textbf{Video Reasoning/Summary} & \textbf{Actual} & \textbf{Predicted} & \textbf{Segment (s)} & \textbf{Target} \\
\midrule

\multirow{4}{*}{\textbf{HateMM}} 
& hate\_video\_354 & The video contains a racist and offensive symbol, text, and image. & Hate & Hate & \makecell{P: 0.17 -- 1.89 \\ GT: 0.0 -- 0.20} & \makecell{P: Blacks, Whites \\ GT: Blacks} \\
\cmidrule(lr){2-7}
& non\_hate\_video\_36 & Two individuals in prison uniforms are engaged in a physical altercation in an outdoor area surrounded by stone walls. & Non Hate & Non Hate & -- & -- \\
\midrule

\multirow{9}{*}{\textbf{MHC}} 
& qFnOYeOmPFQ & The video depicts a discussion among five women where one participant makes a broad statement about all white people being racist. & Hateful & Offensive & \makecell{P: [(0 -- 5)] \\ GT: [(1 -- 26)]} & \makecell{P: White \\ GT: White, LGBTQ} \\
\cmidrule(lr){2-7}
& bVQcoSLoapc & The video shows several men reacting with smiles and laughter, indicating a friendly and humorous interaction. & Offensive & Normal & {GT: [(10 -- 22)]} & {GT: Woman} \\
\cmidrule(lr){2-7}
& TfPv2KiLOrs & A static image of an old Japanese Ukiyo-e style painting with a blurred area and text overlay asking about historical practices. & Normal & Normal & -- & -- \\

\bottomrule
\end{tabular}}
\caption{\textbf{Qualitative Analysis of Model Predictions.} We present five samples across HateMM and MHC datasets, comparing actual labels (GT) with model predictions (P) for classification, timestamp localization, and target identification. For short videos ($<30s$), the reasoning and the summary are the same. However, for long videos with multiple chunks, the summary acts as the context for the next video chunk.}
\label{tab:qualitative_analysis}
\end{table*}

\subsection{Qualitative Analysis}   \label{subsec:qualitative}

Table~\ref{tab:qualitative_analysis} provides qualitative insights into the behavior of the proposed approach on representative samples from HateMM and MultiHateClip (MHC), complementing the quantitative trends observed in Tables~\ref{tab:zs_context_results} and~\ref{tab:tandem_all_datasets}. Due to budget constraints, we are not able to carry out a full human evaluation campaign. Therefore, the authors manually analyzed the model’s output on $10$ test samples of which $5$ are reported in Table~\mbox{\ref{tab:qualitative_analysis}}. In general, these examples illustrate that the primary gains of \acro extend beyond classification accuracy, enabling structured and interpretable predictions that jointly align classification, temporal localization, and target identification. We provide structured reasoning examples in XML format for HateMM and MHC in Appendix~\mbox{\ref{sec:app-xml-samples}}.

On HateMM, the first example highlights a clear success case. The model correctly identifies explicit hate content driven by visual symbols and overlaid text, assigns the correct \emph{Hate} label, and predicts a hate segment that closely overlaps with the ground-truth interval. Although the predicted span is slightly longer than the annotated segment, it remains tightly localized around the relevant content. Importantly, the model correctly identifies both target groups, demonstrating that the learned cross-modal alignment effectively grounds visual cues into structured target predictions. This behavior is consistent with the strong target identification gains achieved by SFT + SCCR + RL in Table~\ref{tab:tandem_all_datasets}. In contrast, the non-hateful HateMM example shows that the model refrains from hallucinating timestamps or targets when none are present, correctly predicting a \emph{Non Hate} label and leaving the temporal and target fields empty. This aligns with the low false-positive behavior implied by the high accuracy and F1 scores in Table~\ref{tab:zs_context_results}, confirming that improvements in target detection do not come at the cost of over-triggering.

The MultiHateClip examples expose more subtle failure modes that help explain the remaining performance gap. In the first MHC sample, the model correctly detects harmful intent but predicts \emph{Offensive} instead of \emph{Hateful}, indicating confusion between closely related classes. The predicted timestamp captures only the initial portion of the hateful utterance, missing later segments of the ground-truth span, and only one of the two annotated targets is identified. This case illustrates the inherent difficulty of multi-class hate categorization and long-range temporal grounding in conversational videos, and mirrors the quantitative observation that macro-F1 and localization scores on MHC remain lower than on HateMM. Nevertheless, the model still produces a partially correct structured output, which would be difficult to obtain using zero-shot inference alone.

The remaining MHC examples demonstrate a conservative prediction behavior. In one case, the model predicts \emph{Normal} for content labeled as \emph{Offensive}, suggesting a tendency to err toward under-classification when contextual cues are ambiguous or when offensiveness is conveyed implicitly through tone rather than explicit language. While this behavior contributes to reduced recall, as reflected in Table~\ref{tab:zs_context_results}, it also helps explain the relatively stable precision observed after SFT and SCCR-based training. The final example represents a clean true negative, where the model correctly classifies non-hateful visual content and avoids producing timestamps or targets, reinforcing that structured predictions remain well-calibrated even in the absence of hate content.

\paragraph{Limitations}
The results on MHC and IHV also highlight dataset-specific challenges that are orthogonal to model capacity. First, in multi-class settings such as MHC and IHV, label imbalance has a significant impact: although overall accuracy remains competitive, macro-F1 is consistently lower (Tables~\ref{tab:zs_context_results},~\ref{tab:tandem_all_datasets}, and~\ref{tab:mhc_binary}), indicating that minority classes are harder to model. This often manifests as conservative predictions biased toward majority labels, even when up/down sampling or class reweighting is applied. Second, the semantic boundary between related labels, such as hateful versus offensive in MHC, or explicit versus implicit hate in IHV, is inherently ambiguous. This ambiguity is exacerbated by real-world audio-visual noise and annotation subjectivity, making it difficult for models to reliably learn fine-grained distinctions. The binary MHC results in Table~\ref{tab:mhc_binary} partially support this observation, as collapsing labels leads to more stable improvements under SFT + SCCR + RL, suggesting that some errors arise from label granularity rather than failure to detect harmful content. Finally, while reinforcement learning has been extensively studied for text-only LLMs, our results indicate that stable RL optimization for VL and AL models remains sensitive to reward design, sampling strategy, and cross-modal conditioning. The performance gaps between SFT-only, RL-only, and SCCR-conditioned RL in Table~\ref{tab:tandem_all_datasets} suggest that best practices for multimodal RL are still emerging, motivating further investigation beyond unimodal settings.

\section{Conclusion and Future Work}    \label{sec:conclusion}

We introduced \acro, a tandem RL framework for structured multimodal hate understanding that jointly optimizes vision-language and audio-language models. By combining supervised fine-tuning with self-constrained context rounds and reward-guided policy optimization, \acro enables consistent improvements across hate classification, temporal localization, and target identification. Extensive experiments on HateMM, MultiHateClip, and ImpliHateVid demonstrate that while strong zero-shot and context-augmented baselines perform competitively on coarse classification, they remain limited in producing temporally grounded and target-aware predictions. In contrast, \acro yields more robust and interpretable structured outputs, narrowing this gap across datasets and achieving $0.73$ F1 in target identification on HateMM.

Our ablation studies highlight three key insights. First, supervised fine-tuning provides essential structural priors but is insufficient for learning fine-grained temporal grounding. Second, reinforcement learning is effective only when constrained by a structured cross-modal context, as provided by SCCR. Third, tandem optimization is critical for preventing modality drift and ensuring consistent cross-modal alignment. Together, these findings validate \acro as a unified framework rather than a collection of independent components.


Several directions remain open for future work. First, replacing string-based inputs/outputs with token identifiers (token-in, token-out) could significantly improve RL stability, particularly for vision-language and audio-language models. Second, exploring classification-preserving training, \mbox{\textit{e.g.}}, a dedicated video-level auxiliary head or adaptive re-weighting of the classification reward, to improve binary hate/non-hate performance while retaining gains in temporal localization and target identification is promising. Third, extending cross-modal context beyond a single video chunk and leveraging larger LoRA parameter budgets may further enhance long-range temporal reasoning with the availability of higher compute resources. Fourth, directly fine-tuning omni models with reinforcement learning remains an open challenge, requiring improved Hugging Face library support and resolution of current technical bottlenecks. Finally, improving dataset quality, particularly by addressing the inherent ambiguity between offensive and hateful labels and increasing coverage of minority classes, remains crucial for reducing bias and improving robustness in multi-class hate understanding.


\section*{Ethical Impact}

This work relies exclusively on publicly available datasets and pre-trained models, and does not involve the collection of new personal data. Nevertheless, like all data-driven systems, the proposed approach may inherit biases present in its training data, potentially leading to uneven performance across under-represented groups or sensitive contexts. The system is intended to support, rather than replace, human moderation by providing structured and interpretable signals such as temporal localization and target identification. Responsible deployment, therefore, requires human oversight, particularly for ambiguous or borderline cases, to mitigate risks of over-censorship, misclassification, or unintended harm. We view this work as a step toward more transparent and accountable moderation tools rather than fully automated decision-making systems.

\bibliography{aaai2026}

\begin{thebibliography}{30}
\providecommand{\natexlab}[1]{#1}

\bibitem[{Bai et~al.(2025)Bai, Chen, Liu, Wang, Ge, Song, Dang, Wang, Wang, Tang et~al.}]{bai2025qwen2}
Bai, S.; Chen, K.; Liu, X.; Wang, J.; Ge, W.; Song, S.; Dang, K.; Wang, P.; Wang, S.; Tang, J.; et~al. 2025.
\newblock Qwen2. 5-vl technical report.
\newblock \emph{arXiv preprint arXiv:2502.13923}.

\bibitem[{Chu et~al.(2024)Chu, Xu, Yang, Wei, Wei, Guo, Leng, Lv, He, Lin et~al.}]{chu2024qwen2}
Chu, Y.; Xu, J.; Yang, Q.; Wei, H.; Wei, X.; Guo, Z.; Leng, Y.; Lv, Y.; He, J.; Lin, J.; et~al. 2024.
\newblock Qwen2-audio technical report.
\newblock \emph{arXiv preprint arXiv:2407.10759}.

\bibitem[{Cinelli et~al.(2021)Cinelli, De~Francisci~Morales, Galeazzi, Quattrociocchi, and Starnini}]{cinelli2021echo}
Cinelli, M.; De~Francisci~Morales, G.; Galeazzi, A.; Quattrociocchi, W.; and Starnini, M. 2021.
\newblock The echo chamber effect on social media.
\newblock \emph{Proceedings of the national academy of sciences}, 118(9): e2023301118.

\bibitem[{Comanici et~al.(2025)Comanici, Bieber, Schaekermann, Pasupat, Sachdeva, Dhillon, Blistein, Ram, Zhang, Rosen et~al.}]{comanici2025gemini}
Comanici, G.; Bieber, E.; Schaekermann, M.; Pasupat, I.; Sachdeva, N.; Dhillon, I.; Blistein, M.; Ram, O.; Zhang, D.; Rosen, E.; et~al. 2025.
\newblock Gemini 2.5: Pushing the frontier with advanced reasoning, multimodality, long context, and next generation agentic capabilities.
\newblock \emph{arXiv preprint arXiv:2507.06261}.

\bibitem[{Das et~al.(2023)Das, Raj, Saha, Mathew, Gupta, and Mukherjee}]{das2023hatemm}
Das, M.; Raj, R.; Saha, P.; Mathew, B.; Gupta, M.; and Mukherjee, A. 2023.
\newblock Hatemm: A multi-modal dataset for hate video classification.
\newblock In \emph{Proceedings of the International AAAI Conference on Web and Social Media}, volume~17, 1014--1023.

\bibitem[{Erickson and Yan(2025)}]{erickson2025content}
Erickson, J.; and Yan, B. 2025.
\newblock Content Moderation and Hate Speech on Alternative Platforms: A Case Study of BitChute.
\newblock \emph{Proceedings of the ACM on Human-Computer Interaction}, 9(2): 1--18.

\bibitem[{Escorcia et~al.(2016)Escorcia, Caba~Heilbron, Niebles, and Ghanem}]{escorcia2016daps}
Escorcia, V.; Caba~Heilbron, F.; Niebles, J.~C.; and Ghanem, B. 2016.
\newblock Daps: Deep action proposals for action understanding.
\newblock In \emph{European conference on computer vision}, 768--784. Springer.

\bibitem[{Guo et~al.(2025)Guo, Liu, Li, Cheng, Tang, Sui, Liu, Chen, and Zhao}]{guo2025vtg}
Guo, Y.; Liu, J.; Li, M.; Cheng, D.; Tang, X.; Sui, D.; Liu, Q.; Chen, X.; and Zhao, K. 2025.
\newblock Vtg-llm: Integrating timestamp knowledge into video llms for enhanced video temporal grounding.
\newblock In \emph{Proceedings of the AAAI Conference on Artificial Intelligence}, volume~39, 3302--3310.

\bibitem[{Guo et~al.(2024)Guo, Liu, Li, Liu, Chen, and Tang}]{guo2024trace}
Guo, Y.; Liu, J.; Li, M.; Liu, Q.; Chen, X.; and Tang, X. 2024.
\newblock Trace: Temporal grounding video llm via causal event modeling.
\newblock \emph{arXiv preprint arXiv:2410.05643}.

\bibitem[{Hee et~al.(2024)Hee, Sharma, Cao, Nandi, Nakov, Chakraborty, and Lee}]{hee2024recent}
Hee, M.~S.; Sharma, S.; Cao, R.; Nandi, P.; Nakov, P.; Chakraborty, T.; and Lee, R. K.-W. 2024.
\newblock Recent advances in online hate speech moderation: Multimodality and the role of large models.
\newblock \emph{Findings of the Association for Computational Linguistics: EMNLP 2024}, 4407--4419.

\bibitem[{Hu et~al.(2022)Hu, Shen, Wallis, Allen-Zhu, Li, Wang, Wang, Chen et~al.}]{hu2022lora}
Hu, E.~J.; Shen, Y.; Wallis, P.; Allen-Zhu, Z.; Li, Y.; Wang, S.; Wang, L.; Chen, W.; et~al. 2022.
\newblock Lora: Low-rank adaptation of large language models.
\newblock \emph{ICLR}, 1(2): 3.

\bibitem[{Kentmen-Cin(2025)}]{kentmen2025hate}
Kentmen-Cin, C. 2025.
\newblock Hate Speech on Social Media: A Systemic Narrative Review of Political Science Contributions.
\newblock \emph{Social Sciences}, 14(10): 610.

\bibitem[{Koushik, Kanojia, and Treharne(2025)}]{koushik2025towards}
Koushik, G.~A.; Kanojia, D.; and Treharne, H. 2025.
\newblock Towards a Robust Framework for Multimodal Hate Detection: A Study on Video vs. Image-based Content.
\newblock In \emph{Companion Proceedings of the ACM on Web Conference 2025}, 2014--2023.

\bibitem[{Lambert et~al.(2024)Lambert, Morrison, Pyatkin, Huang, Ivison, Brahman, Miranda, Liu, Dziri, Lyu et~al.}]{lambert2024tulu}
Lambert, N.; Morrison, J.; Pyatkin, V.; Huang, S.; Ivison, H.; Brahman, F.; Miranda, L. J.~V.; Liu, A.; Dziri, N.; Lyu, S.; et~al. 2024.
\newblock Tulu 3: Pushing frontiers in open language model post-training.
\newblock \emph{arXiv preprint arXiv:2411.15124}.

\bibitem[{Li et~al.(2025{\natexlab{a}})Li, Yu, Huang, Liu, Liang, Liu, Che, Yu, Boyd-Graber, Mi et~al.}]{li2025self}
Li, Z.; Yu, W.; Huang, C.; Liu, R.; Liang, Z.; Liu, F.; Che, J.; Yu, D.; Boyd-Graber, J.; Mi, H.; et~al. 2025{\natexlab{a}}.
\newblock Self-rewarding vision-language model via reasoning decomposition.
\newblock \emph{arXiv preprint arXiv:2508.19652}.

\bibitem[{Li et~al.(2025{\natexlab{b}})Li, Zhang, Guo, Bennamoun, Boussaid, Dwivedi, Gong, and Ke}]{li2025watch}
Li, Z.; Zhang, X.; Guo, Y.; Bennamoun, M.; Boussaid, F.; Dwivedi, G.; Gong, L.; and Ke, Q. 2025{\natexlab{b}}.
\newblock Watch and Listen: Understanding Audio-Visual-Speech Moments with Multimodal LLM.
\newblock \emph{arXiv preprint arXiv:2505.18110}.

\bibitem[{OpenAI(2023)}]{gpt4v2023}
OpenAI, T. 2023.
\newblock GPT-4V(ision) System Card.
\newblock Technical report.

\bibitem[{Rehman et~al.(2025)Rehman, Bhatnagar, Kabde, Bansal, and Kumar}]{rehman2025implihatevid}
Rehman, M. Z.~U.; Bhatnagar, A.; Kabde, O.; Bansal, S.; and Kumar, N. 2025.
\newblock ImpliHateVid: A benchmark dataset and two-stage contrastive learning framework for implicit hate speech detection in videos.
\newblock In \emph{Proceedings of the 63rd Annual Meeting of the Association for Computational Linguistics (Volume 1: Long Papers)}, 17209--17221.

\bibitem[{Shao et~al.(2024)Shao, Wang, Zhu, Xu, Song, Bi, Zhang, Zhang, Li, Wu et~al.}]{shao2024deepseekmath}
Shao, Z.; Wang, P.; Zhu, Q.; Xu, R.; Song, J.; Bi, X.; Zhang, H.; Zhang, M.; Li, Y.; Wu, Y.; et~al. 2024.
\newblock Deepseekmath: Pushing the limits of mathematical reasoning in open language models.
\newblock \emph{arXiv preprint arXiv:2402.03300}.

\bibitem[{Shou, Wang, and Chang(2016)}]{shou2016temporal}
Shou, Z.; Wang, D.; and Chang, S.-F. 2016.
\newblock Temporal action localization in untrimmed videos via multi-stage cnns.
\newblock In \emph{Proceedings of the IEEE conference on computer vision and pattern recognition}, 1049--1058.

\bibitem[{Sun et~al.(2025)Sun, Chen, Zhang, Zhang, Yue, Jiao, and Fu}]{sun2025multihateloc}
Sun, Q.; Chen, T.; Zhang, Y.; Zhang, Y.; Yue, J.; Jiao, J.; and Fu, Z. 2025.
\newblock MultiHateLoc: Towards Temporal Localisation of Multimodal Hate Content in Online Videos.
\newblock \emph{arXiv preprint arXiv:2512.10408}.

\bibitem[{Takada et~al.(2024)Takada, Suzuki, Takushima, Tanoue, Sato, Kumar, Nishihara, Hori, and Ueki}]{takada2024direct}
Takada, T.; Suzuki, Y.; Takushima, H.; Tanoue, H.; Sato, H.; Kumar, A.; Nishihara, H.; Hori, T.; and Ueki, K. 2024.
\newblock Direct Metric Optimization for Image Captioning through Reward-Weighted Augmented Data Utilization.
\newblock In \emph{Proceedings of the 62nd Annual Meeting of the Association for Computational Linguistics (Volume 1: Long Papers)}, 8333--8346.

\bibitem[{Wang, Wang, and Lee(2025)}]{wang2025hateclipseg}
Wang, H.; Wang, Z.; and Lee, R. K.-W. 2025.
\newblock HateClipSeg: A Segment-Level Annotated Dataset for Fine-Grained Hate Video Detection.
\newblock In \emph{Proceedings of the 33rd ACM International Conference on Multimedia}, 13304--13310.

\bibitem[{Wang et~al.(2024)Wang, Yang, Naseem, and Lee}]{wang2024multihateclip}
Wang, H.; Yang, T.~R.; Naseem, U.; and Lee, R. K.-W. 2024.
\newblock Multihateclip: A multilingual benchmark dataset for hateful video detection on youtube and bilibili.
\newblock In \emph{Proceedings of the 32nd ACM International Conference on Multimedia}, 7493--7502.

\bibitem[{Xing et~al.(2025)Xing, Dong, Zang, Cao, Liang, Huang, Wang, Wu, and Lin}]{xing2025caprl}
Xing, L.; Dong, X.; Zang, Y.; Cao, Y.; Liang, J.; Huang, Q.; Wang, J.; Wu, F.; and Lin, D. 2025.
\newblock Caprl: Stimulating dense image caption capabilities via reinforcement learning.
\newblock \emph{arXiv preprint arXiv:2509.22647}.

\bibitem[{Xu et~al.(2025)Xu, Guo, Hu, Chu, Wang, He, Wang, Shi, He, Zhu, Lv, Wang, Guo, Wang, Ma, Zhang, Zhang, Hao, Guo, Yang, Zhang, Ma, Wei, Bai, Chen, Liu, Wang, Yang, Liu, Ren, Zheng, Men, Zhou, Yu, Yang, Yu, Zhou, and Lin}]{xu2025qwen3omnitechnicalreport}
Xu, J.; Guo, Z.; Hu, H.; Chu, Y.; Wang, X.; He, J.; Wang, Y.; Shi, X.; He, T.; Zhu, X.; Lv, Y.; Wang, Y.; Guo, D.; Wang, H.; Ma, L.; Zhang, P.; Zhang, X.; Hao, H.; Guo, Z.; Yang, B.; Zhang, B.; Ma, Z.; Wei, X.; Bai, S.; Chen, K.; Liu, X.; Wang, P.; Yang, M.; Liu, D.; Ren, X.; Zheng, B.; Men, R.; Zhou, F.; Yu, B.; Yang, J.; Yu, L.; Zhou, J.; and Lin, J. 2025.
\newblock Qwen3-Omni Technical Report.

\bibitem[{Yue et~al.(2025)Yue, Yang, Chen, Jiao, and Fu}]{yue2025multimodal}
Yue, J.; Yang, S.; Chen, T.; Jiao, J.; and Fu, Z. 2025.
\newblock Multimodal Hate Detection Using Dual-Stream Graph Neural Networks.
\newblock \emph{arXiv preprint arXiv:2509.13515}.

\bibitem[{Zannettou et~al.(2018)Zannettou, Caulfield, Blackburn, De~Cristofaro, Sirivianos, Stringhini, and Suarez-Tangil}]{zannettou2018origins}
Zannettou, S.; Caulfield, T.; Blackburn, J.; De~Cristofaro, E.; Sirivianos, M.; Stringhini, G.; and Suarez-Tangil, G. 2018.
\newblock On the origins of memes by means of fringe web communities.
\newblock In \emph{Proceedings of the internet measurement conference 2018}, 188--202.

\bibitem[{Zhao et~al.(2017)Zhao, Xiong, Wang, Wu, Tang, and Lin}]{zhao2017temporal}
Zhao, Y.; Xiong, Y.; Wang, L.; Wu, Z.; Tang, X.; and Lin, D. 2017.
\newblock Temporal action detection with structured segment networks.
\newblock In \emph{Proceedings of the IEEE international conference on computer vision}, 2914--2923.

\bibitem[{Zheng et~al.(2025)Zheng, Liu, Li, Chen, Yu, Gao, Dang, Liu, Men, Yang et~al.}]{zheng2025group}
Zheng, C.; Liu, S.; Li, M.; Chen, X.-H.; Yu, B.; Gao, C.; Dang, K.; Liu, Y.; Men, R.; Yang, A.; et~al. 2025.
\newblock Group sequence policy optimization.
\newblock \emph{arXiv preprint arXiv:2507.18071}.

\end{thebibliography}

\section*{Paper Checklist}

\begin{enumerate}

\item For most authors...
\begin{enumerate}
    \item  Would answering this research question advance science without violating social contracts, such as violating privacy norms, perpetuating unfair profiling, exacerbating the socio-economic divide, or implying disrespect to societies or cultures?
    \answerTODO{Yes}
  \item Do your main claims in the abstract and introduction accurately reflect the paper's contributions and scope?
    \answerTODO{Yes}
   \item Do you clarify how the proposed methodological approach is appropriate for the claims made? 
    \answerTODO{Yes}
   \item Do you clarify what are possible artifacts in the data used, given population-specific distributions?
    \answerTODO{NA}
  \item Did you describe the limitations of your work?
    \answerTODO{Yes}
  \item Did you discuss any potential negative societal impacts of your work?
    \answerTODO{Yes}
      \item Did you discuss any potential misuse of your work?
    \answerTODO{Yes}
    \item Did you describe steps taken to prevent or mitigate potential negative outcomes of the research, such as data and model documentation, data anonymization, responsible release, access control, and the reproducibility of findings?
    \answerTODO{Yes}
  \item Have you read the ethics review guidelines and ensured that your paper conforms to them?
    \answerTODO{Yes}
\end{enumerate}

\item Additionally, if your study involves hypotheses testing...
\begin{enumerate}
  \item Did you clearly state the assumptions underlying all theoretical results?
    \answerTODO{NA}
  \item Have you provided justifications for all theoretical results?
    \answerTODO{NA}
  \item Did you discuss competing hypotheses or theories that might challenge or complement your theoretical results?
    \answerTODO{NA}
  \item Have you considered alternative mechanisms or explanations that might account for the same outcomes observed in your study?
    \answerTODO{NA}
  \item Did you address potential biases or limitations in your theoretical framework?
    \answerTODO{NA}
  \item Have you related your theoretical results to the existing literature in social science?
    \answerTODO{NA}
  \item Did you discuss the implications of your theoretical results for policy, practice, or further research in the social science domain?
    \answerTODO{NA}
\end{enumerate}

\item Additionally, if you are including theoretical proofs...
\begin{enumerate}
  \item Did you state the full set of assumptions of all theoretical results?
    \answerTODO{NA}
	\item Did you include complete proofs of all theoretical results?
    \answerTODO{NA}
\end{enumerate}

\item Additionally, if you ran machine learning experiments...
\begin{enumerate}
  \item Did you include the code, data, and instructions needed to reproduce the main experimental results (either in the supplemental material or as a URL)?
    \answerTODO{No. It will be released upon paper's acceptance.}
  \item Did you specify all the training details (e.g., data splits, hyperparameters, how they were chosen)?
    \answerTODO{Yes}
     \item Did you report error bars (e.g., with respect to the random seed after running experiments multiple times)?
    \answerTODO{Yes}
	\item Did you include the total amount of compute and the type of resources used (e.g., type of GPUs, internal cluster, or cloud provider)?
    \answerTODO{Yes}
     \item Do you justify how the proposed evaluation is sufficient and appropriate to the claims made? 
    \answerTODO{Yes}
     \item Do you discuss what is ``the cost`` of misclassification and fault (in)tolerance?
    \answerTODO{NA}
  
\end{enumerate}

\item Additionally, if you are using existing assets (e.g., code, data, models) or curating/releasing new assets, \textbf{without compromising anonymity}...
\begin{enumerate}
  \item If your work uses existing assets, did you cite the creators?
    \answerTODO{Yes}
  \item Did you mention the license of the assets?
    \answerTODO{NA}
  \item Did you include any new assets in the supplemental material or as a URL?
    \answerTODO{No}
  \item Did you discuss whether and how consent was obtained from people whose data you're using/curating?
    \answerTODO{NA}
  \item Did you discuss whether the data you are using/curating contains personally identifiable information or offensive content?
    \answerTODO{NA}
\item If you are curating or releasing new datasets, did you discuss how you intend to make your datasets FAIR?
\answerTODO{NA}
\item If you are curating or releasing new datasets, did you create a Datasheet for the Dataset? 
\answerTODO{NA}
\end{enumerate}

\item Additionally, if you used crowdsourcing or conducted research with human subjects, \textbf{without compromising anonymity}...
\begin{enumerate}
  \item Did you include the full text of instructions given to participants and screenshots?
    \answerTODO{NA}
  \item Did you describe any potential participant risks, with mentions of Institutional Review Board (IRB) approvals?
    \answerTODO{NA}
  \item Did you include the estimated hourly wage paid to participants and the total amount spent on participant compensation?
    \answerTODO{NA}
   \item Did you discuss how data is stored, shared, and deidentified?
   \answerTODO{NA}
\end{enumerate}

\end{enumerate}

\appendix

\section{Implementation Details}    \label{sec:app-imp}

\paragraph{Hyperparameters and Training.} 
All fine-tuning was performed using LoRA with a rank of $8$ and alpha of $16$. We used a learning rate of $5e-5$ and a batch size of $2$ per GPU. LLM generation max sequence length was capped at $384$ tokens with $4$ generations per sample. We utilized the debiased GRPO loss function from the trl\footnote{https://huggingface.co/docs/trl/index} library with a KL penalty of $0$. The cross-modal context history was limited to one chunk due to computational constraints. Experiments were conducted on dual NVIDIA A100 (80GB) GPUs for $200$ steps, requiring approximately $6$ days for HateMM and $72$ hours for MHC.

\paragraph{SFT Data Selection.}
To initialize models with high-quality silver labels, we generated candidate annotations for a subset of $100$ videos using Qwen3-Omni-30B-A3B-Thinking. We strictly filtered these against ground truth, retaining only samples where the model correctly predicted classification labels to serve as structural priors for the SFT phase.

\paragraph{Evaluation Logic.}
For temporal localization, if a model predicts multiple segments, we compute the IoU for each against all ground-truth intervals and assign the maximum overlap score. Segments spanning the $30$-second chunk boundary are treated as truncated intervals. For non-hateful samples, predicting ``None'' or empty sets for targets/timestamps is counted as a correct prediction.

\section{XML Structured Reasoning Samples} \label{sec:app-xml-samples}

We include two structured XML examples from trained \acro predictions, aligned with the qualitative cases discussed in Table~\ref{tab:qualitative_analysis}.

\vspace{0.3cm}
\textbf{Sample i. (HateMM: hate\_video\_354)}
\begin{lstlisting}[language=XML,basicstyle=\scriptsize\ttfamily,breaklines=true]
<sample dataset="HateMM" video_id="hate_video_354">
  <ground_truth>
    <label>Hate</label>
    <segments>
      <segment start="0.00" end="0.20" />
    </segments>
    <targets>
      <target>Blacks</target>
    </targets>
  </ground_truth>
  <prediction>
    <reasoning>
      The video contains a racist and offensive symbol, text, and image.
    </reasoning>
    <label>Hate</label>
    <segments>
      <segment start="0.17" end="1.89" />
    </segments>
    <targets>
      <target>Blacks</target>
      <target>Whites</target>
    </targets>
    <summary>
      Explicit hateful content detected with minority-group targeting.
    </summary>
  </prediction>
</sample>
\end{lstlisting}

\vspace{0.3cm}
\textbf{Sample ii. (HateMM: hate\_video\_299)}
\begin{lstlisting}[language=XML,basicstyle=\scriptsize\ttfamily,breaklines=true]
<sample dataset="HateMM" video_id="hate_video_299">
  <ground_truth>
    <label>Hate</label>
  </ground_truth>
  <prediction>
    <reasoning>
      A conversation between two individuals, one expressing frustration and disinterest in answering questions.
    </reasoning>
    <label>Non-Hate</label>
    <segments></segments>
    <targets></targets>
    <summary>
      No explicit hateful content detected.
    </summary>
  </prediction>
</sample>
\end{lstlisting}

\vspace{0.3cm}
\textbf{Sample iii. (MHC: qFnOYeOmPFQ)}
\begin{lstlisting}[language=XML,basicstyle=\scriptsize\ttfamily,breaklines=true]
<sample dataset="MHC" video_id="qFnOYeOmPFQ">
  <ground_truth>
    <label>Hateful</label>
    <segments>
      <segment start="1.00" end="26.00" />
    </segments>
    <targets>
      <target>White</target>
      <target>LGBTQ</target>
    </targets>
  </ground_truth>
  <prediction>
  <reasoning>
      The video depicts a discussion among five women where one participant makes a broad statement about all white people being racist.
    </reasoning>
    <label>Offensive</label>
    <segments>
      <segment start="0.00" end="5.00" />
    </segments>
    <targets>
      <target>White</target>
    </targets>
    <summary>
      Offensive speech detected against white people.
    </summary>
  </prediction>
</sample>
\end{lstlisting}

\vspace{0.3cm}
\textbf{Sample iv. (MHC: cXRgVEENkPA)}
\begin{lstlisting}[language=XML,basicstyle=\scriptsize\ttfamily,breaklines=true]
<sample dataset="MHC" video_id="cXRgVEENkPA">
  <ground_truth>
    <label>Hateful</label>
    <segments>
      <segment start="0.00" end="47.00" />
    </segments>
    <targets>
      <target>LGBTQ</target>
      <target>Catholics</target>
      <target>Christian</target>
    </targets>
  </ground_truth>
  <prediction>
    <reasoning>
      The video contains strong language and controversial opinions.
    </reasoning>
    <label>Offensive</label>
    <segments></segments>
    <targets></targets>
    <summary>
      Contains strong opinions against christians and LGBTQ people.
    </summary>
  </prediction>
</sample>
\end{lstlisting}

\section{Why classification-only baselines do not constitute directly comparable upper bounds}
\label{app:coarsening}

Let \(X\) denote an input video, \(B \in \{0,1\}\) its binary hate/non-hate label, \(S\) the temporal hate span(s), and \(Z\) the target label/set. Define the structured output as
\[
Y = (B,S,Z).
\]
The key observation is that binary classification is a deterministic projection of the structured task: there exists a map \(g\) such that
\[
B = g(Y).
\]
Therefore, the structured task is a refinement of the binary task.

\paragraph{Proposition.}
Let \(R=f(X)\) be any learned representation. If \(R\) is sufficient for the structured task \(Y\), \textit{i.e.},
\[
p(Y \mid X,R) = p(Y \mid R),
\]
then \(R\) is also sufficient for the binary task \(B\), \textit{i.e.},
\[
p(B \mid X,R) = p(B \mid R).
\]
The converse does not hold in general.

\paragraph{Proof.}
Since \(B=g(Y)\), for any \(b \in \{0,1\}\),
\[
p(B=b \mid X,R)
= \sum_{y:g(y)=b} p(Y=y \mid X,R).
\]
By sufficiency of \(R\) for \(Y\),
\[
\sum_{y:g(y)=b} p(Y=y \mid X,R)
= \sum_{y:g(y)=b} p(Y=y \mid R)
= p(B=b \mid R).
\]
Hence,
\[
p(B=b \mid X,R) = p(B=b \mid R),
\]
which proves that any representation sufficient for predicting \((B,S,Z)\) is automatically sufficient for predicting \(B\). However, the converse fails whenever two videos share the same binary label but differ in their hateful span(s) or target(s). In that case, a representation that preserves only the binary label is sufficient for classification but not for structured prediction. Thus, binary classification is a strict coarsening of the joint task.
\vspace{0.3cm}

\paragraph{Corollary (effect of video length).}
Suppose a video is discretized into \(T\) temporal bins and, for simplicity, each hateful video contains one contiguous hateful span and one target choice from a family of \(M\) admissible target labels/sets. Then the number of distinct positive structured outputs is at least
\[
M \cdot \frac{T(T+1)}{2},
\]
because there are \(T(T+1)/2\) possible start--end intervals. Including the non-hate case gives
\[
|\mathcal{Y}| \ge 1 + M \cdot \frac{T(T+1)}{2}.
\]
By contrast, binary classification has only
\[
|\mathcal{B}| = 2
\]
possible outputs. Therefore, the worst-case description length of the structured output is at least
\[
\log_2\!\left(1 + M \cdot \frac{T(T+1)}{2}\right)
\]
bits, compared with \(1\) bit for binary classification. The gap increases with video length \(T\), and would be even larger if multiple disjoint hateful spans are allowed.

\paragraph{Implication.}
This result does not imply a fixed numerical drop in binary classification accuracy. Rather, it formalizes why classification-only baselines solve a strictly easier and coarser problem than joint classification, temporal localization, and target identification. Under fixed model capacity, supervision, and optimization budget, some trade-off between coarse classification performance and structured understanding is therefore expected.

\end{document}